\documentclass[10pt,twocolumn,letterpaper]{article}

\usepackage[pagenumbers]{cvpr} 

\definecolor{cvprblue}{rgb}{0.21,0.49,0.74}
\usepackage[pagebackref,breaklinks,colorlinks,allcolors=cvprblue]{hyperref}
\usepackage[ruled,vlined,linesnumbered]{algorithm2e}
\usepackage{subcaption}
\usepackage{booktabs}   
\usepackage{multirow}   
\usepackage{graphicx}   
\usepackage[table]{xcolor}  
\usepackage{comment}
\SetKwInOut{Input}{Input}
\SetKwInOut{Output}{Output}
\def\confName{CVPR}
\def\confYear{2027}

\title{Seeing Beyond the Obvious: Complementary Semantic Guidance for Salient Single-Positive Multi-Label Learning}

\author{Tong Jin\thanks{The authors contribute equally.}, 
Meng-Long Wang\footnotemark[1], 
Xingyu Zhao\thanks{Corresponding Author.}, 
Sheng-Jun Huang\footnotemark[2]\\
Nanjing University of Aeronautics and Astronautics\\
Nanjing, China\\
}

\begin{document}
\maketitle
\begin{abstract}
Salient Single-Positive Multi-Label Learning (SalSPML) assigns each image its most salient positive label, leaving other classes unlabeled. Existing methods construct class-level Salient Prototypes from observed positives to guide the recognition of unobserved classes. However, since salient annotations favor visually prominent instances, these prototypes may not fully capture the visual diversity of hidden positives, resulting in insufficient positive supervision.
To address this limitation, we propose \textbf{CoST}, a \textbf{Co}verage-guided Complementary \textbf{S}oft-\textbf{T}arget Refinement framework that supplements Salient Prototypes with complementary class-semantic information. 
Specifically, CoST identifies semantically relevant candidates from unknown classes by leveraging task-learned category information and image-text semantic correspondence. The selected candidates are then aggregated into class-wise Coverage Prototypes to complement the semantic representations provided by Salient Prototypes.
Furthermore, we introduce a ranking-based selective refinement strategy that identifies classes receiving stronger semantic support from Coverage Prototypes and selectively enhances their soft targets while preserving the base soft targets for the rest.
Experiments on multiple benchmarks demonstrate the effectiveness of CoST, with strong performance gains over existing baselines.
\end{abstract}

\section{Introduction}

Multi-Label Learning (MLL) aims to identify multiple semantic categories associated with an image and has been widely studied in various computer vision tasks \cite{li2026prototype, zhou2026rethinking}. However, obtaining exhaustive annotations for all relevant categories is often labor-intensive and costly. To alleviate this burden, Single-Positive Multi-Label Learning (SPML) has emerged as a practical learning setting, where each training image is provided with only one observed positive label while the remaining classes are left unlabeled \cite{cole2021multi}. In conventional SPML settings, the observed positive label is typically selected randomly from positive labels, with the remaining annotations discarded \cite{xie2022label,liu2023revisiting,chen2024boosting}. Although this strategy reduces annotation costs, random label selection may not fully reflect real-world annotation behavior.

In practical annotation scenarios, human annotators tend to focus on the most visually prominent category in an image rather than selecting a positive label randomly. Motivated by this observation, Wang et al.~\cite{wangone} introduced Salient SPML (SalSPML), where the most salient positive category is retained as the only observed label. To exploit the semantic information provided by salient annotations, they further developed a prototype-guided learning framework that constructs class-wise Salient Prototypes from observed positive examples. These prototypes provide class-level semantic references for recognizing potentially overlooked positive categories.

\begin{figure*}[t]
    \centering
    \includegraphics[width=\textwidth]{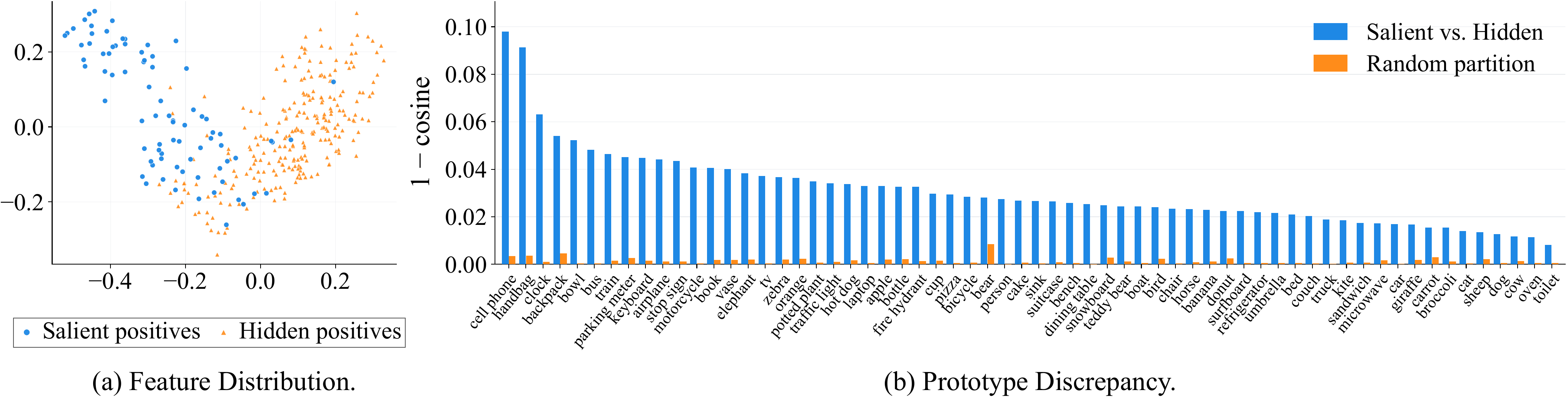}
    \caption{Analysis of semantic discrepancies between salient and hidden positives on MS-COCO. (a) PCA visualization of class-wise embeddings for salient and hidden positives in the handbag category. (b) Class-wise prototype discrepancies, measured by cosine dissimilarity, under the actual Salient-Hidden partition and matched random partitions.}
    \label{fig:motivation}
\end{figure*}

Despite the effectiveness of Salient Prototypes, constructing class representations solely from observed salient positives may not fully capture the semantic diversity of the corresponding classes. This limitation arises because saliency-based annotations favor visually prominent instances, whereas hidden positives may exhibit different visual characteristics due to variations in scale, appearance, and context. Consequently, hidden positives may exhibit weaker similarity to their corresponding Salient Prototypes and receive insufficient positive supervision.
To investigate this issue, we conduct feature distribution and prototype discrepancy analyses on the MS-COCO dataset \cite{lin2014microsoft}, as illustrated in Fig.~\ref{fig:motivation}. The PCA visualization in Fig.~\ref{fig:motivation}(a) reveals a noticeable distribution shift between salient and hidden handbag positives. Furthermore, Fig.~\ref{fig:motivation}(b) shows that the actual Salient-Hidden partition generally exhibits larger prototype discrepancies than matched random partitions. These observations suggest that Salient Prototypes may provide incomplete semantic coverage, limiting their ability to represent the diverse visual appearances of hidden positives (see Appendix~A for more detailed analyses).

A natural way to address this limitation is to introduce complementary class-semantic information from unknown classes. However, unknown classes may correspond to either hidden positives or true negatives. Directly incorporating their features into class representations may therefore introduce unreliable semantic associations and contaminate the resulting prototypes. Consequently, identifying informative candidates from unknown classes and exploiting their complementary semantic information without indiscriminate aggregation are essential to improving the supervision of hidden positives.

To address these challenges, we propose \textbf{CoST}, a \textbf{Co}verage-guided Complementary \textbf{S}oft-\textbf{T}arget Refinement framework for SalSPML.
Specifically, CoST identifies informative candidates from unknown classes through Teacher-guided candidate proposal and CLIP-based semantic verification. Following class-wise selection, the retained candidates are aggregated into class-wise Coverage Prototypes, which complement the semantic representations captured by Salient Prototypes.
Furthermore, we introduce a ranking-based selective refinement strategy that compares the class rankings of the two prototype branches. It selectively enhances the soft targets of classes receiving stronger semantic support from Coverage Prototypes, while preserving the base targets of other unknown classes. 
This enables CoST to exploit complementary semantic evidence without indiscriminately modifying soft targets.

The main contributions are summarized as follows:

\begin{itemize}
\item We investigate semantic differences between salient positives and hidden positives, revealing the limited coverage of Salient Prototypes in capturing diverse appearances.
\item We develop a complementary prototype construction strategy that leverages task-specific class-discriminative knowledge and external vision-language semantics to identify unknown-class candidates. The resulting Coverage Prototypes provide complementary semantic representations that extend the coverage of Salient Prototypes.
\item Building on base soft-target supervision, we develop a ranking-based selective refinement strategy to identify complementary classes and selectively enhance their soft targets while preserving the base targets for other unknown classes. This provides additional semantic supervision for potentially overlooked positives.
\item Extensive experiments on multiple benchmarks demonstrate the effectiveness of CoST compared with baselines.
\end{itemize}

\section{Related Work}

\subsection{Multi-Label Learning}
Multi-Label Learning (MLL) aims to predict multiple semantic categories associated with each instance and has been widely studied in visual recognition tasks~\cite{zhang2013review,pan2023miml,kang2025beyond}. Unlike single-label classification, MLL requires models to capture both instance-label associations and inter-label dependencies~\cite{zhao2021transformer,chen2024pursuit,gao2025multi}. However, obtaining exhaustive annotations is often labor-intensive and costly, motivating research on MLL with incomplete labels~\cite{hu2023dualcoop,kim2023bridging,si2026revisiting}. In this setting, only a subset of labels is observed, while the remaining labels may correspond to either positive or negative categories. Treating all unobserved labels as negatives can introduce false-negative supervision. To address this issue, existing methods exploit label correlations \cite{sun2021global,sun2026weighted}, identify potential positives \cite{zhang2023learning,lyu2025addressing}, and incorporate semantic information \cite{ding2023exploring,chen2026adapting} to alleviate incomplete supervision.

\subsection{Single-Positive Multi-Label Learning}

Single-Positive Multi-Label Learning (SPML) is an extreme incomplete-supervision setting where each image retains only one observed positive label, with all remaining labels treated as unknown. 
Cole et al.~\cite{cole2021multi} first systematically investigated this problem, demonstrating that effective multi-label classifiers can be trained using only one positive label per image. Traditional SPML typically randomly selects one positive label from the complete label set of each image. 
Subsequent methods improve unknown-label recognition through pseudo-label estimation \cite{cole2021multi,xu2022one,zhou2022acknowledging,liu2023revisiting}, robust loss design and false-negative identification~\cite{chen2024boosting,song2024ignore,tran2025more}, auxiliary label-cardinality information~\cite{gharib2025single,liu2026can}, and vision-language model-based semantic guidance~\cite{wang2023hierarchical,xing2024vision}.

However, real-world annotations tend to favor visually prominent categories rather than random selection. To reflect this behavior, Wang et al.~\cite{wangone} introduced Salient SPML (SalSPML), retaining the most salient positive category per image. They further proposed PiSA, which constructs class-wise Salient Prototypes from observed positives and employs prototype-guided embedding learning to recognize hidden positives. 
Although Salient Prototypes provide reliable class-level semantic guidance, their reliance on observed salient positives may limit their ability to capture the diverse visual appearances of hidden positives. Therefore, incorporating complementary class-semantic information beyond Salient Prototypes remains important for improving hidden-positive supervision.

\section{Methodology}
\label{sec:methodology}

\subsection{Preliminaries}
\label{sec:preliminaries}

We consider Salient Single-Positive Multi-Label Learning (SalSPML)~\cite{wangone}, in which each image is annotated with a single salient positive label. Formally, the training set is defined as $\mathcal{D}=\{(x_i,y_i^{\mathrm{sal}})\}_{i=1}^{N}$, where $x_i$ represents the $i$-th image and $y_i^{\mathrm{sal}}\in\{1,\ldots,C\}$ specifies its observed salient positive class. Here, $N$ and $C$ denote the numbers of examples and classes, respectively.

Unlike fully supervised Multi-Label Learning (MLL), SalSPML provides no annotations for the other classes. Accordingly, the unknown label set of image $x_i$ is defined as $\mathcal{U}_i = \{1,\ldots,C\} \setminus \{y_i^{\mathrm{sal}}\}$. These unknown classes may contain either unannotated positive labels or true negative labels, which cannot be directly distinguished from the available annotations. Treating all unknown classes as negatives may suppress the learning of hidden positives, whereas assigning excessive positive supervision may introduce incorrect class associations \cite{cole2021multi}.
To address this ambiguity, we adopt a soft-target learning framework that provides continuous supervision for unknown classes.

Let $\ell_{ic}$ denote the logit predicted for class $c$ in image $x_i$, with the corresponding probability given by $p_{ic}=\sigma(\ell_{ic})$, where $\sigma(\cdot)$ denotes the sigmoid function. 
Since the observed salient class is known to be positive, we use the standard positive binary cross-entropy term to define the positive classification loss:
\begin{equation}
    \mathcal{L}_{\mathrm{pos}}
    = -\frac{1}{N}\sum_{i=1}^{N}
    \log p_{i,y_i^{\mathrm{sal}}}.
\end{equation}

For each unknown class $c\in\mathcal{U}_i$, we assign a soft target $t_{ic}\in[0,1]$ and define the soft binary cross-entropy loss as
\begin{align}
\mathcal{L}_{\mathrm{unk}}
&= -\frac{1}{N(C-1)}
\sum_{i=1}^{N}\sum_{c\in\mathcal{U}_i}
\Big[
t_{ic}\log p_{ic} \label{eq:unk}\\
&\qquad +(1-t_{ic})\log(1-p_{ic})
\Big],
\notag
\end{align}
where $C-1$ normalizes the loss across unknown classes per image. Unlike binary pseudo-labels, soft targets provide graded supervision by balancing positive and negative learning signals. A higher $t_{ic}$ encourages a larger predicted probability for class $c$, while a lower value favors its suppression. 
The overall classification objective is defined as
\begin{equation}
    \mathcal{L}_{\mathrm{cls}}
    = \mathcal{L}_{\mathrm{pos}}
    + \eta\mathcal{L}_{\mathrm{unk}},
\label{eq:cls}
\end{equation}
where $\eta$ weights unknown-class supervision.

To preserve the reliable class-level semantic information captured by Salient Prototypes from observed positive examples, we retain the prototype learning loss $\mathcal{L}_{\mathrm{PL}}$ from PiSA~\cite{wangone}.
The overall training objective is then given by
\begin{equation}
    \mathcal{L}
    = \mathcal{L}_{\mathrm{cls}}
    + \alpha_{\mathrm{PL}}\mathcal{L}_{\mathrm{PL}},
\label{eq:total}
\end{equation}
where $\alpha_{\mathrm{PL}}$ controls the contribution of the prototype loss.

The effectiveness of soft-target learning largely depends on the quality of the targets assigned to unknown classes. Although Salient Prototypes provide reliable class-level semantic guidance, their reliance on observed salient positive samples may limit their ability to capture the diverse visual appearances of each class. Consequently, hidden positives with less salient or atypical visual characteristics may exhibit low similarity to their corresponding prototypes, resulting in insufficient positive supervision.

\begin{figure*}[t]
    \centering
    \includegraphics[width=0.915\textwidth]{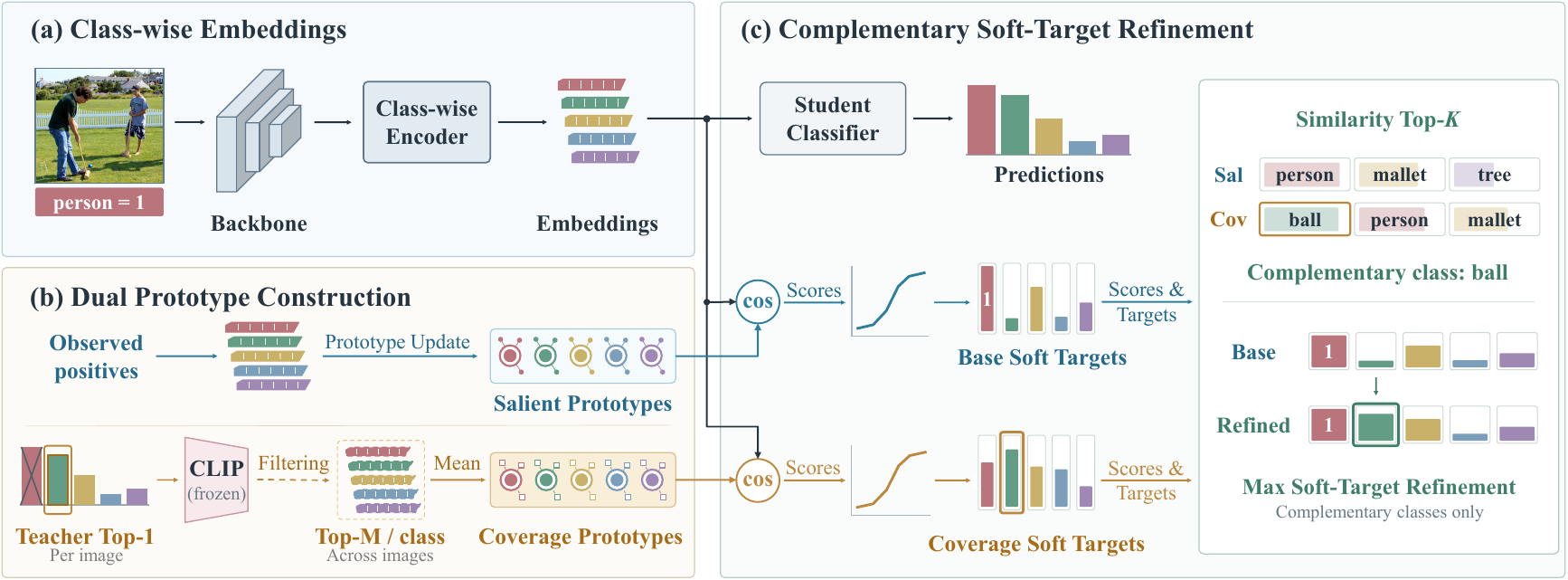}
\caption{Overview of the proposed CoST framework.
(a) Class-wise embeddings are extracted from input images.
(b) Salient Prototypes are constructed from observed salient positives, while Coverage Prototypes are built from Teacher-proposed candidates through CLIP-based semantic verification and class-wise selection.
(c) The two prototype branches independently generate calibrated soft targets, which are selectively refined based on complementary classes identified through similarity-based Top-$K$ rankings.}
\label{fig:cost_framework}
\end{figure*}

To address this limitation, we propose \textbf{CoST} (\textbf{Co}verage-guided Complementary \textbf{S}oft-\textbf{T}arget Refinement). As illustrated in \cref{fig:cost_framework}, CoST constructs Coverage Prototypes from selected unknown-class candidates to complement the semantic information captured by Salient Prototypes. By identifying classes better supported by Coverage Prototypes, CoST selectively refines their soft targets to enhance supervision for potentially overlooked positives.

\subsection{Salient-Guided Base Soft-Target Learning}
\label{sec:salient}

We first construct Salient Prototypes from observed salient positives to establish base supervision for unknown classes~\cite{wangone}. Given an image $x_i$, the backbone and class-wise encoder extract an embedding $z_{ic}\in\mathbb{R}^d$ for each class $c$, where $d$ denotes the feature dimension. These embeddings are then used to construct and update the corresponding Salient Prototype $P_c^{\mathrm{Sal}}\in\mathbb{R}^d$ using examples whose observed salient label is $c$. The prototypes provide class-level semantic references grounded in observed positive labels.

To generate soft targets for unknown classes, we measure the similarity between each class-wise embedding and its corresponding Salient Prototype using cosine similarity:
\begin{equation}
    s_{ic}^{\mathrm{Sal}}
    = \frac{z_{ic}^{\top}P_c^{\mathrm{Sal}}}
    {\|z_{ic}\|_2\|P_c^{\mathrm{Sal}}\|_2}.
\end{equation}

A higher similarity indicates stronger semantic alignment with the corresponding Salient Prototype, providing a basis for assigning soft supervision to unknown classes.

Since similarity distributions may vary across classes, we adopt class-wise threshold calibration. For each class $c$, the threshold is determined by the $q$-quantile of its similarity scores over training samples where $c$ is unobserved:
\begin{equation}
\tau_c^{\mathrm{Sal}}
=
Q_q
\left(
\left\{
s_{ic}^{\mathrm{Sal}}
\mid i=1,\ldots,N,\ c\in\mathcal{U}_i
\right\}
\right),
    \label{eq:sal_threshold}
\end{equation}
where $Q_q(\cdot)$ denotes the quantile function and $q$ specifies the quantile level.
The calibrated similarity is then transformed into a continuous soft target:
\begin{equation}
t_{ic}^{\mathrm{Sal}}
=
\sigma\left(
\frac{s_{ic}^{\mathrm{Sal}}-\tau_c^{\mathrm{Sal}}}
{T}
\right),
    \quad c\in\mathcal{U}_i,
    \label{eq:sal_soft_target}
\end{equation}
where $T>0$ is a temperature parameter controlling the sensitivity of soft targets to similarity differences, and the resulting soft targets are detached.

The resulting $t_{ic}^{\mathrm{Sal}}$ serves as the base soft target for unknown classes, while the observed salient positive class retains its target of $1$. Since Salient Prototypes are derived solely from observed salient positives, we further construct Coverage Prototypes from selected unknown-class candidates to provide complementary semantic references.

\subsection{Coverage Prototype Construction}
\label{sec:coverage}

Since unknown classes may correspond to either hidden positives or true negatives, directly aggregating their features may introduce unreliable class associations. We therefore adopt a progressive candidate selection strategy consisting of Teacher-guided candidate proposal, CLIP-based semantic verification, and class-wise candidate selection.

\textbf{Teacher-Guided Candidate Proposal.}
In SalSPML, the observed label represents the most salient positive, while other hidden positives, though less salient than the observed category, may still exhibit distinctive visual cues and receive relatively high model confidence. We therefore leverage learned class-discriminative knowledge to identify promising candidates among unknown classes.
To obtain stable candidate proposals, we adopt an epoch-wise Teacher derived from the model at the end of the preceding epoch. The Teacher provides class-wise predictions for candidate proposals and the corresponding class-wise embeddings for Coverage Prototype construction.

Let $p_{ic}^{\mathrm{Tea}}$ denote the probability predicted by the Teacher for class $c$ on image $x_i$. For each image, we exclude its observed salient positive class and select the unknown class with the highest Teacher confidence:
\begin{equation}
    c_i^*
    =\arg\max_{c\in\mathcal{U}_i}
    p_{ic}^{\mathrm{Tea}}.
    \label{eq:teacher_top1}
\end{equation}

Accordingly, each image contributes at most one candidate class, forming the initial candidate set
\begin{equation}
    \mathcal{A}_i = \{c_i^*\}.
\end{equation}

This Top-1 strategy prioritizes the most confident unknown-class prediction while limiting the number of candidates contributed by each image. The selected class serves only as a candidate proposal rather than a direct positive label, since the Teacher is trained under incomplete supervision and may produce incorrect class associations, motivating the subsequent semantic filtering.

\textbf{CLIP-Based Semantic Verification.}
To provide additional semantic evidence for candidate filtering, we introduce a pretrained vision-language model, CLIP \cite{radford2021learning}, to evaluate the semantic compatibility between images and their corresponding class descriptions. Let $f_I(\cdot)$ and $f_T(\cdot)$ denote the image and text encoders of CLIP, respectively. Given an image $x_i$ and a class-specific text description $d_c$, constructed using the prompt template \textit{``a photo of a \{class\}''}, we extract their normalized representations:

\begin{equation}
    v_i=
    \frac{f_I(x_i)}{\|f_I(x_i)\|_2},
    \qquad
    e_c=
    \frac{f_T(d_c)}{\|f_T(d_c)\|_2}.
\end{equation}

The image-text similarity is then computed as
\begin{equation}
    r_{ic}^{\mathrm{CLIP}} = v_i^\top e_c.
    \label{eq:clip_similarity}
\end{equation}

To avoid manually specifying similarity cutoffs, we adopt class-wise thresholds, each derived from CLIP similarities of training images with observed salient label $c$:
\begin{equation}
    \tau_c^{\mathrm{CLIP}}
    =
    Q_{q_{\mathrm{CLIP}}}
    \left(
    \left\{
    r_{ic}^{\mathrm{CLIP}}
    \mid y_i^{\mathrm{sal}}=c
    \right\}_{i=1}^{N}
    \right),
    \label{eq:clip_threshold}
\end{equation}
where $Q_{q_{\mathrm{CLIP}}}(\cdot)$ denotes the quantile function and $q_{\mathrm{CLIP}}$ specifies the quantile level.

For each proposed candidate $c_i^*$, we retain it only when its image-text similarity meets the corresponding class-specific threshold:
\begin{equation}
    g_i=
    \mathbb{I}\left[
    r_{i,c_i^*}^{\mathrm{CLIP}}
    \geq \tau_{c_i^*}^{\mathrm{CLIP}}
    \right],
    \label{eq:clip_filter}
\end{equation}
where $\mathbb{I}[\cdot]$ denotes the indicator function.

The resulting candidate set is
\begin{equation}
    \widetilde{\mathcal{A}}_i =
    \begin{cases}
        \{c_i^*\}, & g_i=1,\\
        \varnothing, & g_i=0.
    \end{cases}
\end{equation}

The Teacher and CLIP serve complementary roles in candidate selection: the former identifies candidates based on task-specific class-discriminative knowledge, while the latter evaluates their image-text semantic compatibility.
Passing the CLIP-based criterion provides additional semantic support for a candidate, but does not necessarily indicate that it is a true hidden positive.

\textbf{Class-wise Candidate Aggregation.}
After image-level candidate filtering, we aggregate the retained candidates according to their proposed classes. For each class $c$, the corresponding candidate set is defined as
\begin{equation}
    \widetilde{\mathcal{C}}_c
    =
    \left\{
    i \mid c\in\widetilde{\mathcal{A}}_i
    \right\}.
\end{equation}

Since the number of candidates may vary substantially across classes and some retained candidates may still contain incorrect class associations, directly aggregating all candidates may introduce unnecessary noise. We therefore further rank the candidates within each class according to their Teacher confidence and retain at most $M$ candidates:
\begin{equation}
    \mathcal{C}_c
    =
    \operatorname{Top}_{M}
    \left(
    \widetilde{\mathcal{C}}_c;
    \{p_{ic}^{\mathrm{Tea}}\}_{i\in\widetilde{\mathcal{C}}_c}
    \right),
    \label{eq:class_topm}
\end{equation}
where $\operatorname{Top}_{M}$ returns the indices of at most $M$ candidates with the highest Teacher confidence, and $M$ denotes the maximum number of candidates retained for each class. If fewer than $M$ valid candidates are available, all are retained.

Finally, we construct a Coverage Prototype for each class with at least one selected candidate by averaging their Teacher embeddings:
\begin{equation}
    P_c^{\mathrm{Cov}}
    =
    \frac{1}{|\mathcal{C}_c|}
    \sum_{i\in\mathcal{C}_c}
    z_{ic}^{\mathrm{Tea}},
    \qquad |\mathcal{C}_c|>0,
    \label{eq:coverage_prototype}
\end{equation}
where $z_{ic}^{\mathrm{Tea}}\in\mathbb{R}^d$ denotes the class-wise embedding of image $x_i$ for class $c$, extracted by the Teacher. Accordingly, we define the set of classes with Coverage Prototypes as
\begin{equation}
    \mathcal{V}^{\mathrm{Cov}}
    =
    \{c\in\{1,\ldots,C\}:|\mathcal{C}_c|>0\}.
    \label{eq:valid_cov_classes}
\end{equation}

Unlike Salient Prototypes, which are constructed from samples with observed salient positive labels, Coverage Prototypes are derived from selected unknown-class candidates. The two types of prototypes are represented in the same $d$-dimensional class-wise embedding space but are constructed from different sources of supervision. They therefore provide complementary class-semantic references for the subsequent soft-target refinement (see Appendix~B for theoretical analysis).

\subsection{Complementary Soft-Target Refinement}
\label{sec:refinement}

After constructing Salient and Coverage Prototypes, CoST leverages their complementary semantic information to refine the base soft targets for unknown classes. We compute prototype similarities using the current Student embeddings $z_{ic}$, while Coverage Prototypes serve as references constructed from the preceding-epoch Teacher embeddings. Since the two types of prototypes are constructed from different sources, they may provide different levels of semantic support for the same class. Rather than directly combining all soft targets, we identify classes for which Coverage Prototypes provide additional semantic support and selectively enhance their corresponding base soft targets.

\textbf{Coverage-Guided Soft-Target Generation.}
Given the constructed Coverage Prototypes, we first measure the cosine similarity between the current Student embedding $z_{ic}$ and its corresponding Coverage Prototype $P_c^{\mathrm{Cov}}$:
\begin{equation}
    s_{ic}^{\mathrm{Cov}}
    =
    \frac{
    z_{ic}^{\top}P_c^{\mathrm{Cov}}
    }{
    \|z_{ic}\|_2
    \|P_c^{\mathrm{Cov}}\|_2
    }.
    \label{eq:cov_similarity}
\end{equation}

Since similarity distributions may vary across classes, we independently calibrate the similarity scores for each class using a class-specific quantile threshold:
\begin{equation}
\tau_c^{\mathrm{Cov}}
=
Q_q
\left(
\left\{
s_{ic}^{\mathrm{Cov}}
\mid i=1,\ldots,N,\ c\in\mathcal{U}_i
\right\}
\right),
 c\in\mathcal{V}^{\mathrm{Cov}},
    \label{eq:cov_threshold}
\end{equation}
using the same quantile level $q$ as in the Salient branch. The calibrated similarities are then transformed into continuous soft targets:
\begin{equation}
    t_{ic}^{\mathrm{Cov}}
    =
    \sigma\left(
    \frac{
    s_{ic}^{\mathrm{Cov}}-\tau_c^{\mathrm{Cov}}
    }{
    T
    }
    \right),
    \quad c\in\mathcal{U}_i\cap\mathcal{V}^{\mathrm{Cov}},
    \label{eq:cov_soft_target}
\end{equation}
using the same temperature $T$ as in the Salient branch. Together with the Salient soft targets $t_{ic}^{\mathrm{Sal}}$, these targets provide two independently calibrated sources of semantic supervision for unknown classes.

\textbf{Ranking-Based Complementary Class Identification.}
Although Coverage Prototypes provide additional semantic references, their contribution may vary across classes. Applying Coverage-based supervision to all unknown classes may therefore introduce unreliable semantic associations. To identify classes that receive additional semantic support from the Coverage branch, we compare the similarity rankings of the two prototype branches.

Specifically, for each image, we rank classes according to their raw prototype similarities and select the top-$K$ classes from each branch:
\begin{equation}
    \mathcal{T}_i^{\mathrm{Sal}}
    =
    \operatorname{Top}_{K}
    \left(
    \{s_{ic}^{\mathrm{Sal}}\}_{c=1}^{C}
    \right),
    \label{eq:sal_topk}
\end{equation}
\begin{equation}
    \mathcal{T}_i^{\mathrm{Cov}}
    =
    \operatorname{Top}_{K}
    \left(
    \{s_{ic}^{\mathrm{Cov}}\}_{c\in\mathcal{V}^{\mathrm{Cov}}}
    \right),
    \label{eq:cov_topk}
\end{equation}
where $\mathcal{V}^{\mathrm{Cov}}$ denotes the set of classes with available Coverage Prototypes. We use raw similarity scores rather than calibrated soft targets for ranking, since the two branches employ independent class-wise calibration.

The complementary class set is then obtained by taking the difference between the two top-$K$ sets and excluding the observed salient positive class:
\begin{equation}
    \mathcal{M}_i
    =
    \left(
    \mathcal{T}_i^{\mathrm{Cov}}
    \setminus
    \mathcal{T}_i^{\mathrm{Sal}}
    \right)
    \cap
    \mathcal{U}_i.
    \label{eq:complementary_set}
\end{equation}
Here, $\mathcal{M}_i$ contains unknown classes that rank among the top-$K$ under Coverage Prototypes but not under Salient Prototypes. These classes indicate where the Coverage branch provides additional relative semantic support beyond the Salient branch.

\textbf{Selective Soft-Target Enhancement.}
Based on the identified complementary classes, we selectively refine the base soft targets using the Coverage-based supervision. Specifically, we adopt an element-wise maximum operation to enhance the soft targets of complementary classes:
\begin{equation}
    t_{ic}^{*}
    =
    \begin{cases}
        \max\left(
        t_{ic}^{\mathrm{Sal}},
        t_{ic}^{\mathrm{Cov}}
        \right),
        & c\in\mathcal{M}_i,\\[4pt]
        t_{ic}^{\mathrm{Sal}},
        & c\in\mathcal{U}_i
        \setminus\mathcal{M}_i.
    \end{cases}
    \label{eq:refined_soft_target}
\end{equation}

For complementary classes, the maximum operation allows the Coverage branch to strengthen the base supervision when it provides a higher soft target, while preserving the original Salient-based target otherwise. For all remaining unknown classes, the original soft targets are retained without modification. The target of the observed salient positive class remains unchanged at $1$.

The refined soft targets $t_{ic}^{*}$ are used in the unknown-class classification loss defined in \cref{eq:unk}, replacing the base soft targets $t_{ic}$. This selective enhancement enables CoST to incorporate complementary class-semantic information while preserving the supervision provided by Salient Prototypes.

\begin{table*}[t]
\centering
\caption{Performance comparison (mAP, \%) under three SalSPML annotation settings. The best results are highlighted in bold.}
\label{tab:main_results}
\setlength{\tabcolsep}{4pt}
\renewcommand{\arraystretch}{1.1}
\scriptsize
\resizebox{\textwidth}{!}{
\begin{tabular}{lcccccccc}
\toprule
\multirow{2}{*}{Method}
& \multicolumn{3}{c}{Confidence-based}
& \multicolumn{2}{c}{Size-based}
& \multicolumn{3}{c}{CLIP-based} \\
\cmidrule(lr){2-4}\cmidrule(lr){5-6}\cmidrule(lr){7-9}
& VOC & COCO & NUS & VOC & COCO & VOC & COCO & NUS \\
\midrule

Full
& 91.73$\pm$0.04 & 79.18$\pm$0.18 & 51.52$\pm$0.13
& 91.73$\pm$0.04 & 79.18$\pm$0.18
& 91.73$\pm$0.04 & 79.18$\pm$0.18 & 51.52$\pm$0.13 \\

\midrule

BCE
& 83.24$\pm$0.21 & 56.27$\pm$0.16 & 31.66$\pm$0.43
& 64.83$\pm$0.46 & 46.05$\pm$0.17
& 83.63$\pm$0.21 & 62.70$\pm$0.07 & 44.13$\pm$0.09 \\

ASL
& 87.99$\pm$0.15 & 59.34$\pm$0.21 & 34.03$\pm$0.15
& 82.23$\pm$0.13 & 47.76$\pm$0.08
& 86.98$\pm$0.26 & 64.31$\pm$0.01 & 45.27$\pm$0.14 \\

LL-R
& 87.69$\pm$0.12 & 63.65$\pm$0.50 & 33.01$\pm$0.27
& 76.61$\pm$0.47 & 51.61$\pm$0.18
& 87.41$\pm$0.14 & 66.47$\pm$0.13 & 46.43$\pm$0.27 \\

LL-Ct
& 84.60$\pm$0.76 & 62.41$\pm$0.10 & 33.40$\pm$0.32
& 74.20$\pm$0.40 & 51.14$\pm$0.20
& 83.66$\pm$0.07 & 65.07$\pm$0.14 & 46.14$\pm$0.21 \\

BoostLU
& 87.88$\pm$0.15 & 64.53$\pm$0.38 & 33.12$\pm$0.24
& 75.31$\pm$1.03 & 52.37$\pm$0.21
& 87.65$\pm$0.02 & 67.29$\pm$0.19 & 45.78$\pm$0.03 \\

ROLE
& 87.95$\pm$0.26 & 57.83$\pm$0.21 & 31.76$\pm$0.03
& 82.09$\pm$0.16 & 47.36$\pm$0.15
& 87.37$\pm$0.32 & 63.32$\pm$0.36 & 44.62$\pm$0.14 \\

EM
& 89.00$\pm$0.14 & 61.44$\pm$0.04 & 34.76$\pm$0.32
& 83.10$\pm$0.20 & 49.50$\pm$0.07
& 87.97$\pm$0.01 & 65.88$\pm$0.24 & 45.65$\pm$0.39 \\

GR-Loss
& 89.10$\pm$0.06 & 63.58$\pm$0.37 & 36.23$\pm$0.14
& 84.21$\pm$0.19 & 52.05$\pm$0.18
& 87.98$\pm$0.08 & 67.48$\pm$0.13 & \textbf{47.47$\pm$0.16} \\

CS
& 87.96$\pm$0.18 & 62.20$\pm$0.29 & 35.69$\pm$0.24
& 81.22$\pm$0.30 & 50.61$\pm$0.31
& 87.01$\pm$0.31 & 66.20$\pm$0.19 & 47.27$\pm$0.09 \\

GPR-Loss
& 89.61$\pm$0.12 & 67.52$\pm$0.12 & 36.21$\pm$0.05
& 80.94$\pm$0.35 & 54.52$\pm$0.23
& 89.12$\pm$0.19 & 68.47$\pm$0.38 & 47.36$\pm$0.13 \\

PiSA
& 90.21$\pm$0.28 & 66.99$\pm$0.05 & 35.17$\pm$0.15
& 86.38$\pm$0.05 & 54.11$\pm$0.09
& 89.04$\pm$0.31 & 69.72$\pm$0.33 & 46.34$\pm$0.02 \\

\midrule
\rowcolor[gray]{0.92}
\textbf{CoST}
& \textbf{90.36$\pm$0.23}
& \textbf{69.75$\pm$0.01}
& \textbf{37.07$\pm$0.61}
& \textbf{87.07$\pm$0.21}
& \textbf{58.42$\pm$0.26}
&\textbf{89.71$\pm$0.15} & \textbf{70.80$\pm$0.13} & 46.62$\pm$0.13 \\

\bottomrule
\end{tabular}
}
\end{table*}

\section{Experiments}
\subsection{Experimental Setup}

\textbf{Data Preparation.} Following the synthetic data construction protocol in \cite{wangone}, we construct SalSPML benchmarks from fully annotated MS-COCO (COCO) \cite{lin2014microsoft}, Pascal VOC (VOC) \cite{everingham2010pascal}, and NUS-WIDE (NUS) \cite{chua2009nus}. We consider three selection strategies: confidence-based, retaining the positive class with the highest prediction confidence; size-based, selecting the class with the largest annotated region (COCO and VOC); and CLIP-based, retaining the positive class with the highest CLIP confidence to simulate human saliency judgments. Under each strategy, only the selected positive label is retained, while others are treated as unknown. We also evaluate random selection, with results provided in Appendix D.2.
We further evaluate different supervision levels on Visual Genome~\cite{krishna2017visual} by varying the global label-retention rate under confidence-based selection.

\textbf{Comparison Methods.} We compare CoST with representative baselines across three categories. Standard Multi-Label Learning (MLL) methods include BCE and ASL \cite{ridnik2021asymmetric}, while partial-label MLL methods comprise LL-R, LL-Ct \cite{kim2022large}, and BoostLU \cite{kim2023bridging}. For single-positive MLL, we consider ROLE \cite{cole2021multi}, EM \cite{zhou2022acknowledging}, GR-Loss \cite{chen2024boosting}, CS \cite{gharib2025single}, GPR-Loss \cite{tran2025more}, and PiSA \cite{wangone}.

\textbf{Implementation Details.} We adopt an ImageNet \cite{deng2009imagenet}-pretrained ResNet-50 \cite{he2016deep} backbone with a one-layer class-wise attention decoder that produces 2,048-dimensional embeddings. Images are resized to $448\times448$. All models are optimized using Adam \cite{kingma2014adam} with a learning rate of $1\times10^{-5}$, a batch size of 16, and weight decay of $5\times10^{-5}$. We use a one-epoch warm-up with the AN loss \cite{cole2021multi}. Training lasts 20 epochs on VOC and 10 epochs on other datasets. 
A frozen CLIP ViT-B/32 \cite{radford2021learning} is used for semantic verification and CLIP-based selection.
We set $\eta=0.5$ in \cref{eq:cls} and $\alpha_{\mathrm{PL}}=1$ in \cref{eq:total}. For soft-target generation, we set $q=0.975$ in \cref{eq:sal_threshold,eq:cov_threshold} and $T=0.02$ in \cref{eq:sal_soft_target,eq:cov_soft_target}. We further set $q_{\mathrm{CLIP}}=0.4$ in \cref{eq:clip_threshold}, $M=100$ in \cref{eq:class_topm}, and $K=3$ in \cref{eq:sal_topk,eq:cov_topk}.
We reserve 20\% of the training data for validation, with both validation and test sets retaining complete multi-label annotations. The checkpoint with the highest validation mAP is selected for final evaluation. We report test mAP as the mean $\pm$ standard deviation over three runs with different random seeds.

\subsection{Performance Comparison}

We compare CoST with representative baselines under different annotation strategies. As shown in \cref{tab:main_results}, CoST achieves strong performance across datasets and annotation settings, with a 3.90-point gain over the strongest baseline on size-based COCO. Under CLIP-based annotation, CoST remains competitive on NUS, slightly trailing GR-Loss but still surpassing PiSA, possibly reflecting the challenges posed by a larger label space and stronger inter-class correlations. These results demonstrate the effectiveness of CoST in improving hidden-positive recognition.

To further evaluate CoST under varying supervision levels, we conduct experiments on Visual Genome with different positive-label retention rates (30\%, 50\%, and 70\%). As shown in \cref{tab:vg_retention}, CoST consistently outperforms existing methods across all retention rates, demonstrating its ability to leverage limited positive supervision effectively.

\begin{table}[t]
\centering
\caption{Performance comparison under different positive-label retention rates on the Visual Genome dataset.}
\label{tab:vg_retention}
\setlength{\tabcolsep}{8pt}
\renewcommand{\arraystretch}{1.1}
\small
\resizebox{\linewidth}{!}{
\begin{tabular}{lccc}
\toprule
\multirow{2}{*}{Method}
& \multicolumn{3}{c}{Retention Rate} \\
\cmidrule(lr){2-4}
& 30\% & 50\% & 70\% \\
\midrule
BCE
& 43.09$\pm$0.09 & 45.26$\pm$0.01 & 46.54$\pm$0.06 \\

ASL
& 43.44$\pm$0.04 & 45.93$\pm$0.09 & 47.46$\pm$0.04 \\

LL-R
& 44.57$\pm$0.02 & 46.94$\pm$0.02 & 48.61$\pm$0.08 \\

LL-Ct
& 44.18$\pm$0.09 & 46.73$\pm$0.06 & 48.43$\pm$0.06 \\

EM
& 42.47$\pm$0.01 & 45.29$\pm$0.02 & 46.78$\pm$0.11 \\

ML-Decoder
& 44.00$\pm$0.02 & 47.15$\pm$0.02 & 49.24$\pm$0.04 \\

BoostLU
& 45.11$\pm$0.07 & 47.44$\pm$0.06 & 49.17$\pm$0.04 \\

PiSA
& 45.66$\pm$0.07 & 48.24$\pm$0.03 & 49.64$\pm$0.09 \\

\midrule
\rowcolor[gray]{0.92}
\textbf{CoST} &\textbf{46.67$\pm$0.06} & \textbf{48.88$\pm$0.09} & \textbf{49.65$\pm$0.10} \\
\bottomrule
\end{tabular}
}
\end{table}

\subsection{Ablation analysis}

\begin{table}[t]
\centering

\caption{Ablation study of key components in CoST (single seed).
\textit{w/o Coverage} removes the Coverage branch;
\textit{w/o CLIP} removes CLIP-based candidate verification;
\textit{w/o Top-$M$} retains all verified candidates without class-wise selection;
\textit{w/o RankDiff} applies Max enhancement to all Coverage-supported classes.}
\label{tab:ablation}
\setlength{\tabcolsep}{7pt}
\renewcommand{\arraystretch}{1.1}
\small
\resizebox{\linewidth}{!}{
\begin{tabular}{lccccc}
\toprule
\multirow{2}{*}{Variant}
& \multicolumn{3}{c}{Confidence-based}
& \multicolumn{2}{c}{Size-based} \\
\cmidrule(lr){2-4}\cmidrule(lr){5-6}
& VOC & COCO & NUS & VOC & COCO \\
\midrule
w/o Coverage
& 90.21 & 68.55 & 36.63 &86.84 & 57.32 \\
w/o CLIP
& 90.39 & 68.64 & 37.57 & 87.02 & 58.66 \\
w/o Top-M
& 90.28 & 68.98 & 37.61 & 86.98 & 58.47 \\
w/o RankDiff
& 90.49 & 69.54 & 37.12 & 87.00 & 58.15 \\
\midrule
\textbf{CoST}
& 90.61 & 69.76 & 37.77 & 87.23 & 58.77 \\
\bottomrule
\end{tabular}
}
\end{table}

\textbf{Effect of Individual Components.}
We conduct ablation experiments to evaluate the contributions of components in CoST. 
As shown in \cref{tab:ablation}, removing the Coverage branch leads to notable performance drops, particularly on COCO, with decreases of 1.21 and 2.01 mAP points under confidence-based and size-based annotations, respectively. Removing CLIP verification or Top-$M$ selection also degrades performance, while replacing selective refinement with uniform Max enhancement consistently reduces accuracy across the reported settings. These results validate the contributions of all components to CoST.

\begin{figure}[t]
    \centering
    \includegraphics[width=\linewidth]{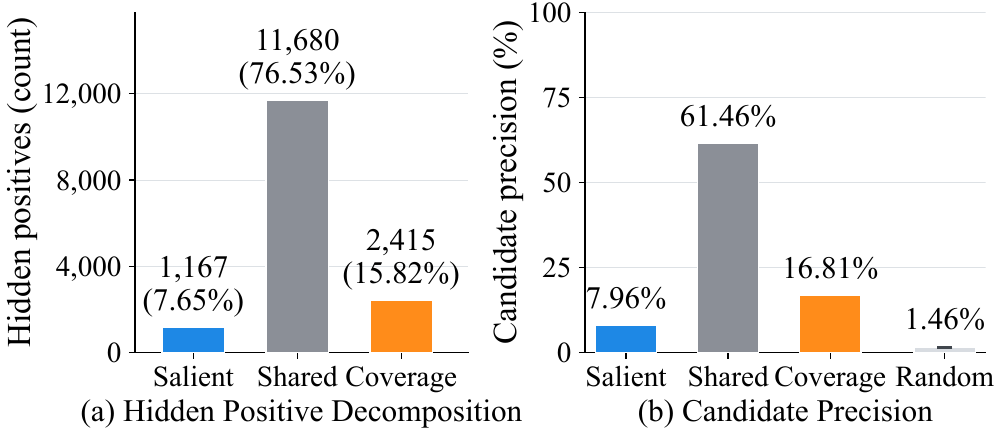}
    \caption{Semantic complementarity between Salient and Coverage representations on COCO.
    (a) Decomposition of hidden positives identified exclusively by Salient, jointly by both branches, or exclusively by Coverage.
    (b) Candidate precision of the subsets, including a budget-matched random complement baseline.}
    \label{fig:semantic_complementarity}
\end{figure}

\textbf{Semantic Complementarity Analysis.}
We analyze the complementarity between Salient and Coverage representations on COCO. As shown in Fig.~\ref{fig:semantic_complementarity}(a), Coverage uniquely identifies 2,415 hidden positives, 18.80\% more than Salient alone. Fig.~\ref{fig:semantic_complementarity}(b) shows that Coverage-only candidates achieve higher precision than the random baseline (16.81\% vs. 1.46\%), while shared candidates exhibit the highest reliability. These results demonstrate the complementary value of Coverage and motivate selective soft-target refinement.

\begin{figure}[t]
\centering
\includegraphics[width=\linewidth]{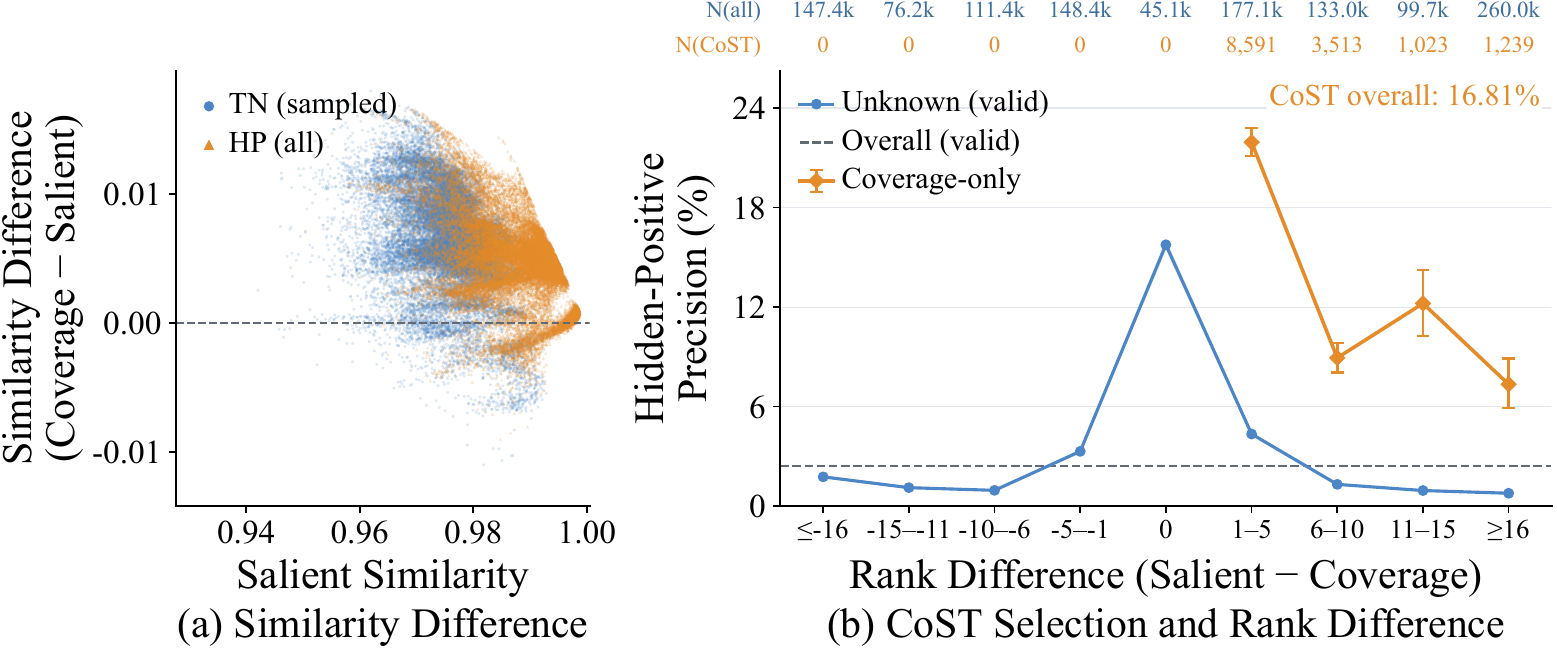}
\caption{Analysis of CoST selection on COCO. (a) Similarity difference between Coverage and Salient representations for hidden positives and sampled true negatives. (b) Hidden-positive precision across ranking-difference intervals, comparing all valid unknown pairs with CoST-selected candidates.  Positive rank differences indicate a better (lower) Coverage rank.}
\label{fig:cost_selection_rank_analysis}
\end{figure}

\textbf{Ranking-based Selection Analysis.}
Fig.~\ref{fig:cost_selection_rank_analysis} illustrates the complementary evidence from Coverage. As shown in Fig.~\ref{fig:cost_selection_rank_analysis}(a), hidden positives exhibit diverse inter-branch similarity differences, making raw similarity insufficient for selection. 
Fig.~\ref{fig:cost_selection_rank_analysis}(b) shows candidates with better Coverage ranks are more likely to be hidden positives. CoST achieves 16.81\% precision, surpassing the valid unknown-pair baseline and supporting ranking-based selective refinement.

\begin{figure}[t]
\centering
\includegraphics[width=\linewidth]{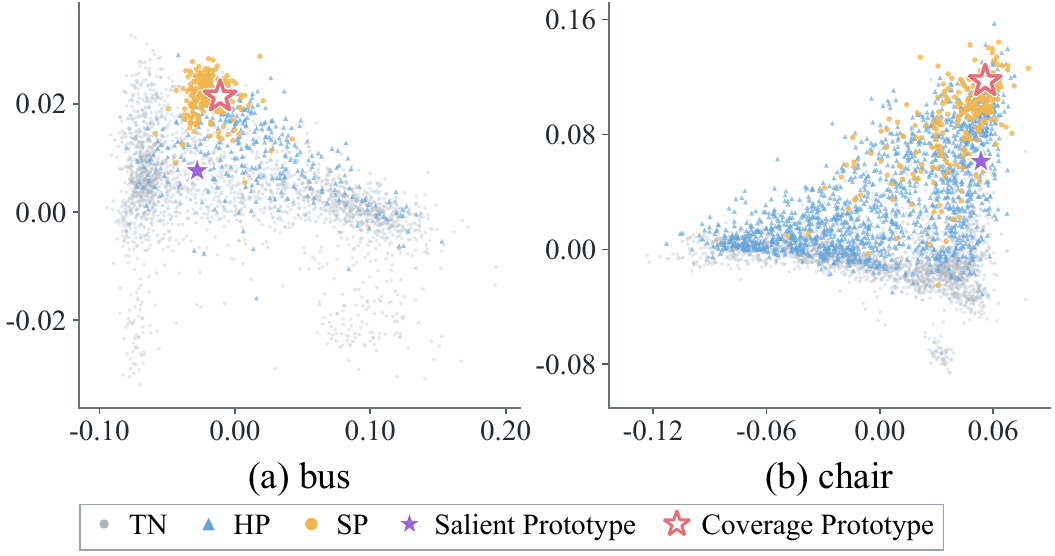}
\caption{PCA visualization of Salient and Coverage prototypes on COCO. SP, HP, and TN denote salient positives, hidden positives, and true negatives, respectively. Stars indicate the two prototypes.}
\label{fig:prototype_visualization}
\end{figure}

\textbf{Prototype Space Visualization.}
Fig.~\ref{fig:prototype_visualization} visualizes the spatial relationships between prototypes and sample groups on COCO. In both categories, prototypes occupy distinct locations, while hidden positives exhibit broader distributions than salient positives. This indicates that Coverage provides complementary semantic references, while its overlap with true negatives motivates selective refinement instead of directly treating similarity as positive confidence.

\begin{figure}[t]
    \centering
    \includegraphics[width=0.89\linewidth]{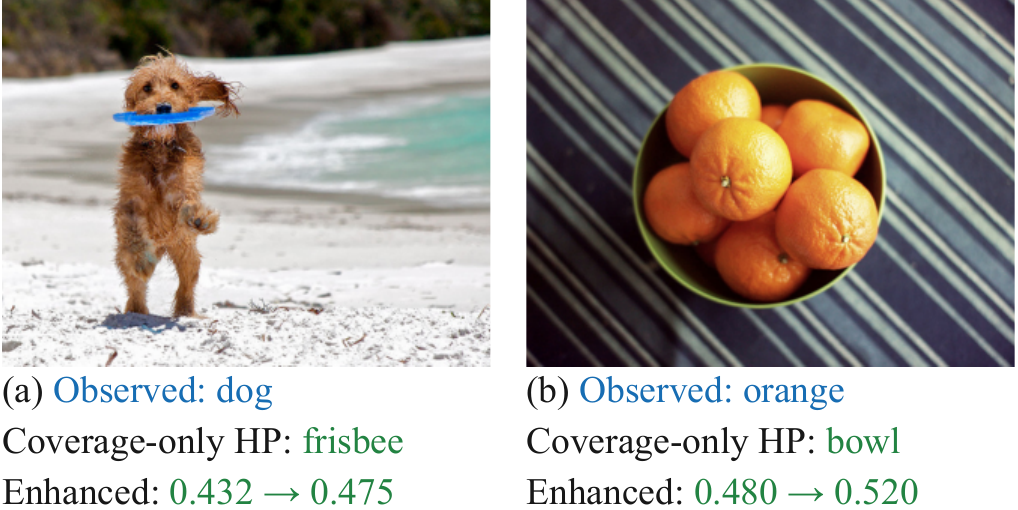}
    \caption{Examples of CoST soft-target refinement. Blue and green denote salient labels and Coverage-identified hidden positives, respectively; values indicate target changes.}
    \label{fig:qualitative_examples}
\end{figure}

\textbf{Qualitative Analysis.}
Fig.~\ref{fig:qualitative_examples} presents two examples on COCO, where Coverage identifies hidden positives missed by Salient Top-$K$, such as \textit{frisbee} and \textit{bowl}. CoST selectively enhances their soft targets, demonstrating the value of representations for learning hidden positives.

\begin{figure}[t]
    \centering
    \includegraphics[width=\linewidth]{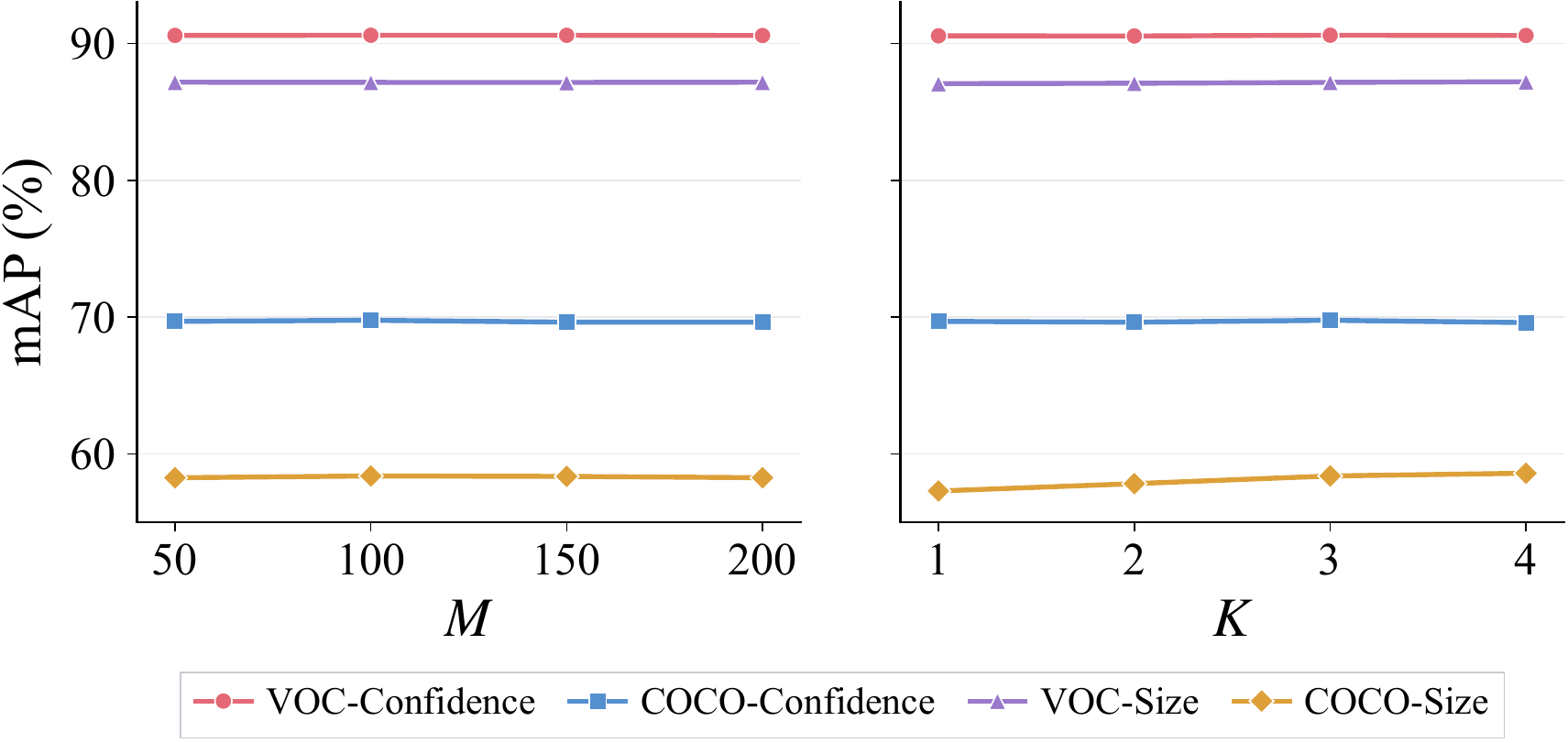}
    \caption{Sensitivity analysis of $M$ and $K$.}
    \label{fig:sensitivity}
\end{figure}

\textbf{Sensitivity Analysis.}
We evaluate the sensitivity of CoST to $M$ and $K$. As shown in \cref{fig:sensitivity}, CoST remains stable across different values.
Further ablation and hyperparameter analyses are provided in Appendix~D.3.


\section{Conclusion}
In this work, we propose the CoST framework for SalSPML. CoST constructs Coverage Prototypes from selected unknown-class candidates to complement semantic information captured by Salient Prototypes. Furthermore, a ranking-based strategy selectively enhances soft targets of complementary classes while preserving base supervision for other unknown classes. Extensive experiments on multiple benchmarks demonstrate the effectiveness of CoST, with strong performance gains over baselines.

\section*{LLM Usage Statement}
We used OpenAI Codex for writing assistance, including language polishing, grammar correction, text condensation, and consistency checks on technical descriptions. This assistance concerned the main text, figure captions, and supplementary material. 

{
    \small
    \bibliographystyle{ieeenat_fullname}
    \bibliography{main}
}


\end{document}